\documentclass[12pt,oneside,reqno]{amsart}

\usepackage{graphicx}
\usepackage{pict2e}
\usepackage{amssymb}
\usepackage{amsthm}                
\usepackage[margin=1in]{geometry}  
\usepackage{graphics}
\usepackage{epstopdf}
\usepackage{amsmath}
\usepackage{graphicx}
\usepackage{subfigure}
\usepackage{latexsym}
\usepackage{amsmath}
\usepackage{amsfonts}
\usepackage{amssymb}
\usepackage{tikz}
\usepackage{mathdots}
\usepackage{comment}
\usepackage{float}
\usepackage[mathscr]{euscript}
\usepackage{enumerate}
\usepackage{epsfig}
\usepackage { graphicx }
\usepackage { hyperref }
\usepackage { graphics }
\usepackage { graphicx }

\usepackage{MnSymbol} 
\usepackage{booktabs}
\usepackage{wrapfig,lipsum,booktabs}

\usepackage{caption}

\usepackage{algorithm}

\usepackage{etoolbox}

\usepackage {eufrak}
\usepackage[utf8]{inputenc}
\usepackage{longtable}
\usepackage[lite]{amsrefs}

\theoremstyle{definition}
\newtheorem{theorem}{Theorem}[section]

\theoremstyle{definition}

\newtheorem{example}[theorem]{Example}
\theoremstyle{remark}

\numberwithin{equation}{section}

\usepackage{newunicodechar}
\numberwithin{equation}{section}

\DeclareUnicodeCharacter{00A0}{~}
\DeclareMathOperator{\Mat}{Mat}
\DeclareMathOperator{\Prob}{Prob}

\DeclareMathOperator{\PH}{PH}
\title{Data-Centric AI Requires Rethinking Data Notion}

\author{Mustafa Hajij}
\address{Santa Clara University}
\email{mhajij@scu.edu}

\author{Ghada Zamzmi}
\address{University of South Florida}
\email{ghadh@mail.usf.edu}

\author{Karthikeyan Natesan Ramamurthy}
\address{IBM Research}
\email{knatesa@us.ibm.com}

\author{Aldo Guzman}
\address{IBM Research}
\email{Aldo.Guzman.Saenz@ibm.com}

\date{}

\dedicatory{}

\begin{document}

\maketitle
\vspace{-25pt}
\begin{abstract}
The transition towards data-centric AI requires revisiting data notions from mathematical and implementational standpoints to obtain unified data-centric machine learning packages. Towards this end, we propose unifying principles offered by categorical and cochain notions of data, and discuss the importance of these principles in data-centric AI transition. In the categorical notion, data is viewed as a mathematical structure that we act upon via morphisms to preserve this structure. As for cochain notion, data can be viewed as a function defined in a discrete domain of interest and acted upon via operators. While these notions are almost orthogonal, they provide a unifying definition to view data, ultimately impacting the way machine learning packages are developed, implemented, and utilized by practitioners.




\end{abstract}

\vspace{-15pt}
\section{Introduction}

The common notion of data, which is overwhelmingly preferred by machine learning practitioners, is rooted in statistical science \cite{friedman2001elements}.  Within this notion, data is described as a sample obtained from a probability density function (PDF) defined over a domain of interest\footnote{There are several others notions rooted in physical and social sciences that we do not focus on here.}. Consider, for instance, the dimensionality reduction problem in which one is given a data $X \subset \mathcal{X}$, and the aim is to find a mapping $f:\mathcal{X}\to \mathcal{Y}$ such that some mathematical structure related to $X$ is preserved via $f$; $\mathcal{X}$ and $\mathcal{Y}$ denote the data space and the target space, respectively. While the data $X$ is typically sampled from an unknown probability distribution, from the perspective of the dimensionality reduction algorithm, the interesting aspect of data lies in the mathematical structure the function $f$ is designed to preserve. 

The past decade has witnessed an enormous interest in the mathematical aspects of machine learning models, their properties, limitations, and generalizations. As the interest in the machine learning community is shifting towards a data-centric AI, we believe that the commonly used notion of data must be carefully revisited and systematically studied. Systematic and cohesive mathematical definition of data notions can impact how data is understood and how machine learning packages are implemented. Also, it can ultimately expand the number of practitioners, who can potentially benefit from emerging machine learning technologies. As an example, we give the analogy of the popular scikit-learn package \cite{scikit-learn}, which has a simple API, yet unified across all underlying algorithms.






To provide a concrete definition of data, we distinguish between two main notions: \textit{categorical} and \textit{cochain}. In the \textit{categorical notion} (Section \ref{cat}), the \textit{invariants} of data are often the interesting object of study. Invariants of data are the structural properties of interest that are preserved under a particular type of \textit{morphisms} (structure-preserving maps). For instance, one invariant can be the metric space structure, encoded in the distance matrix or the persistence diagrams, that is invariant under metric preserving functions (morphism). In the \textit{cochain notion} (Section \ref{cochain}), the domain upon which the data is defined is the main element of interest. Within this perspective, we think of data as a function defined on the elementary building blocks of the domain of interest; this function is acted upon by operators defined on these domains. An example of this type is the solution of a partial differential equation on a surface.
While these two notions are almost orthogonal, they  provide a unifying definition to view the data, ultimately impacting the way machine learning packages are designed, implemented, and utilized by various practitioners.

\vspace{-8pt}
\section{Categorical Notion}
\vspace{-8pt}
\label{cat}
Here, we consider data as a mathematical structure (e.g., a metric space) that we act upon via \textit{morphisms} to preserve this structure. Specifically, the categorical setting seeks \textit{invariants} in the data, a well-defined mathematical quantity extracted from the data and is preserved under a particular type of morphisms. In practical settings, a loss function is typically utilized to preserve the structure as much as possible; however, completely preserving the mathematical structure might not be feasible. In what follows, we provide examples of mathematical structures along with our categorical definitions.



\subsection{Data as a Metric Space}
In the multi-dimensional scaling algorithm (MDS) \cite{cox2008multidimensional}, the data $X \subset \mathcal{X}$ is given as a finite metric space $(X,d_X)$. The purpose of the MDS algorithm is to construct a mapping $f: \mathcal{X} \to \mathcal{Y}$ that preserves the metric structure defined on $X$ as much as possible. Specifically, we define a cost function: 
 
 $\mathcal{L}_{MDS} : Obj(\Mat)\times Obj(\Mat) \to \mathbb{R}$ defined via :
 \begin{equation}
     \mathcal{L}_{MDS} (( (X,d_X)  , (f(X),d_Y ) ) ) = \sum_{(x_1,x_2)\in X\times X } l_{MDS}(d_X (x_1,x_2)),d_Y(f(x_1),f(x_2))
 \end{equation}
where $\Mat$ denotes the category\footnote{A category is a collection of 'objects' (e.g., metric spaces and sets) that are linked by 'arrows' (the maps between these spaces) such that (1) arrows are composable in the same way we compose functions and (2) every object has an identity arrow.}
of metric spaces, $Obj(\Mat)$ denotes the objects of $\Mat$, and $l_{MDS}: \mathbb{R}\times \mathbb{R} \to \mathbb{R} $ is an appropriate loss function (e.g., absolute value). In simple terms, the above cost function defines an approximate notion of equivalence in the category of metric spaces.


\subsection{Data as a Probabilistic Structure}
T-SNE algorithm \cite{hinton2002stochastic} considers a matrix $P$ that encodes a \textit{probabilistic structure} extracted from the data $X$; i.e., given a point cloud $X \subset \mathcal{X} $, the entry $P_{ij}$ encodes the similarity of points $x_i$ and $x_j$ in the data $X$. For brevity, we omit the precise definition of the matrix $P$. For T-SNE, the structure of interest to be preserved when mapping this data to another space is encoded in the matrix $P$.

The algorithm seeks morphisms that (approximately) preserve this structure, hence achieving approximate invariance. We realize the T-SNE algorithm in our weak categorical setting. Namely, the objects are tuples of the form $(X,P_X)$, where $X \subset \mathcal{X}$ is a finite set and $P_X$ is a symmetric matrix such that the summation of its entries add up to 1. The morphisms are functions of the form $f: \mathcal{X} \to \mathcal{Y}$ that preserves the probabilistic structure of $X$ as much as possible. We define a cost function:
 
 $\mathcal{L}_{TSNE} : Obj(\Prob)\times Obj(\Prob) \to \mathbb{R}$ defined via :
 \begin{equation}
     \mathcal{L}_{TSNE} (( (X,P_X)  , (f(X),P_{f(X)} ) ) ) = KL(P_X|| P_{f(X)} )
 \end{equation}
where $KL(A||B)$ is the KL divergence between two probability distributions $A$ and $B$.   


\subsection{Data as a Homological Structure}

Given a point cloud $X \subset \mathcal{X}$, persistent homology \cite{edelsbrunner2010computational} provides a mechanism to construct a finite sequence of topological fingerprints of $X$, which we denote by $ \PH_X:=( \mathbf{D}_X= \{D_1,D_2,...\},{\boldsymbol\pi}_X=  \{\pi_1,\pi_2,...\} )$. Here, $D_i$ are the \textit{persistence pairings} and $\pi_i$ are the \textit{persistence diagrams} induced by the Vietoris–Rips of $(X,d_X)$ \footnote{We only discuss persistence diagrams induced by the Vietoris–Rips filtrations, in such case the persistence computation depends only on the distance matrix $d_X$. The treatment for general filtration is essentially similar.}. These topological fingerprints encode a structure about the data that is relevant in many machine learning-related tasks such as classification and dimensionality reduction \cite{moor2020topological,kachan2020persistent}. Consider the scenario where we construct a mapping $f: \mathcal{X} \to \mathcal{Y}$ such that $f$ preserves these fingerprints as much as possible. Preserving this structure is equivalent to preserving particular entries in the distance matrices $d_X$ and $d_{Y}$, and these entries depend on the persistence pairings $\boldsymbol\pi$ \cite{edelsbrunner2010computational}. The weak equivalence is a function of the form $\mathcal{L}_{PH}: Obj(Per)\times Obj(Per) \to \mathbb{R}$ defined by
\begin{equation}
    \mathcal{L}_{PH}( \PH_X,\PH_{f(X)} )= d(\PH_X,\PH_{f(X)})
\end{equation}
where $d$ is some appropriately chosen distance between the persistence structures of $X$ and $f(X)$ (e.g., the Sinkhorn distance) \cite{frogner2015learning}. See \cite{kachan2020persistent} for recent works on preserving the persistence diagram.  

 \label{clfers}


\vspace{-9pt}
\section{Cochain Notion}
\label{cochain}
Orthogonal to the categorical notion, data can be viewed as a function defined in a discrete domain of interest and acted upon via \textit{operators}. Consider, for instance, a discrete domain $\mathcal{M}$, say a graph or a simplicial complex \cite{hatcher2005algebraic}. Roughly speaking, data in this case is defined in the form $f:\mathcal{M}\to \mathcal{X}$, where $f$ associates to every building block $v\in \mathcal{M}$ a data element $f(v) \in \mathcal{X}$; e.g., a scalar or a vector in some Euclidean space. An important example of this setting is Topological Data Analysis (TDA) \cite{edelsbrunner2010computational}, where one studies the topological properties of the domain $\mathcal{M}$ by studying the properties of functions defined in it. Another example is the Graph Signal Processing (GSP) \cite{ortega2018graph}, where one is interested in the scalar signal defined on the node or the edge sets of a given graph. Similar to the categorical notion, cochain notion provides an abstraction that comes with important virtues: (1) it unifies notions in a single cohesive and mathematically elegant perspective, (2) it becomes evident how to create new similar structures, and (3) it impacts the way machine learning packages are designed and implemented. In what follows, we restrict our work to simplicial complexes and assume these complexes are oriented and finite. Moreover, we use $\mathcal{C}_k(\mathcal{M})$ to denote the space of all linear combinations of all the $k$-simplices in $\mathcal{M}$; elements in $\mathcal{C}_k(\mathcal{M})$ are called chains. Dually, $\mathcal{C}^k(\mathcal{M})$ is the vector space spanned by all real-valued functions defined on $k$-simplices, and its elements are called cochains.  


\vspace{-6pt}
\subsection{Deep Learning on Complexes and Discrete Exterior Calculus}
We argue that the cochain perspective provides a unified framework between two subjects, namely topological deep learning on complexes \cite{hajijcell,ebli2020simplicial,bunch2020simplicial} and discrete exterior calculus (DEC) on complexes \cite{hirani2003discrete,de2020discrete,ptavckova2021simple}, that are typically treated separately (see Section 3.1.2). 




\subsubsection{Discrete exterior calculus}
\label{linear}

 In the context of discrete exterior calculus, a linear operator $\mathcal{A} : \mathcal{C}^i(\mathcal{M}) \to \mathcal{C}^j(\mathcal{M}) $, may act on a cochain $f$ to produce another cochain $\mathcal{A}(f)$ :
\begin{equation}
\label{1234}
    f \xrightarrow[]{\mathcal{A}} \mathcal{A} (f)
\end{equation}

An example of an operator $\mathcal{A}$ is the graph Laplacian. Namely, given a graph $G(V,E)$, the graph Laplacian is a matrix $L=A-D$, where $A$ is the adjacency matrix of the graph and $D$ is the degree matrix. As an operator, the graph Laplacian $L:\mathcal{C}^0 \to \mathcal{C}^0$ takes a signal $f:V\to \mathbb{R}$, a scalar function defined on the node set graph $G$, or an element in $\in \mathcal{C}^0$, and produces another signal $L(f) \in \mathcal{C}^0$.

The (primitive) discrete exterior operators are exactly seven operators that include the exterior derivative (coboundary map), the hodge star, and the wedge product.  We refer the reader to \cite{desbrun2008discrete} for a complete list of these operators. In addition to these operators, there are other complex operators that can be built from them. These operators are the analog of the classical differential operators such as the gradient, divergence, and curl operators. Together cochains and the operators that act on them allow a concrete framework that facilitates computing a cochain of interest such as a cochain obtained by solving a partial differential equation on a discrete surface. To keep our treatment concrete, more examples are provided.

\begin{example}
One of the most primitive operators is the (co)boundary maps. Consider a simplicial complex surface $\mathcal{M}$, and denote its sets of nodes, edges and faces by $V,E,F$. The boundary operator $\partial_k: \mathcal{C}_k \to \mathcal{C}_{k-1} $
is defined to be $\partial_k\{v_0v_1,...,v_k\}=\sum_{i=1}^k (-1)^k \{v_0,...,\hat{v}_i,...,v_k\}$, where $\hat{v}_i$ indicates that $v_i$ is missing from the sequence that determines the simplex. Dually, the coboundary maps are defined to be the transpose $\partial_k^T$ of the boundary maps. The coboundary map $\partial_k^T$ is also referred to by the $k^{th}$ discrete exterior derivative and is denoted by $d^k$. In particular, $d^0(f)$, $d^1 (f)$, and $d^2 (f)$ of some 0, 1, 2 cochains defined on $\mathcal{M}$ are the discrete analogs of the gradient $\nabla f$, the curl $ \nabla\times f $, and divergence $ \nabla \cdot f $ of a smooth function defined on a smooth surface.  
\end{example}

\begin{example}
The (co)boundary can be used to define the $k$-Hodge Laplacian $L_k = \partial_k^\top \partial_k + \partial_{k+1}\partial_{k+1}^\top :  $.  As an operator, the $k-$ Hodge Laplacian maps a k-cochain to another k-cochain.
\end{example}



\subsubsection{Topological Nets: deep learning on complexes and a unifying perspective}
To show that the cochain notion provides a unified framework between topological deep learning on complexes and DEC on complexes, we introduce topological networks (TNs).

Let $f$ be a cochain defined on a finite complex $\mathcal{M}$. We denote by $\mathcal{C}(\mathcal{M})$ to the $\mathbb{R}$-vector space of all linear combinations of such maps. Given an ordering on the cells in $\mathcal{M}$, these cells can be identified with the canonical orthonormal basis in $\mathbb{R}^{|\mathcal{M}|}$. We assume this identification in what follows. A linear operator $\mathcal{A}: \mathcal{C}(\mathcal{M})\to \mathcal{C}(\mathcal{M})$ induces a \textit{topological network} $TN_\mathcal{A}:\mathcal{C}(\mathcal{M})\to \mathcal{C}(\mathcal{M})$ defined via
\begin{equation}
\label{123}
    f \xrightarrow[]{TN_\mathcal{A}} \phi(\mathcal{\mathcal{A}} f W)
\end{equation}
where $W$ is a trainable weight matrix and $\phi$ is a non-linearity. The Topological Networks (TNs) given in equation (\ref{123}) are deceptively simple yet they are the essence of our general unifying framework. Observe first that equation (\ref{123}) generalizes the familiar convolutional graph neural network \cite{kipf2016semi}.  Moreover, it is easy to show that the setting given in equation (\ref{123}) is equivalent to a message passing protocol (determined by the matrix $\mathcal{A}$). See \cite{gilmer2017neural} for a proof in a graph neural network, which easily carries over to higher order complexes such as cell and simplicial complexes. More importantly, we view equation (\ref{123}) from the exterior calculus perspective. Specifically, we emphasize the operator and cochain natures of both $\mathcal{A}$ and $f$ respectively since this perspective allows for a unified treatment of deep learning on complexes with exterior calculus. In particular, when the matrix $W$ is the (untrainable) identity matrix and the map $\phi$ is the identity map then equation (\ref{123}) is equivalent to the linear setting given in equation (\ref{1234}) and discussed in Section \ref{linear}. In other words, DEC which is generally summarized by operations of the form given via equation (\ref{123}) corresponds precisely to a message function in the deep learning setting without performing the update function.  

This unified framework offers multiple advantages. First, it is designed to express and train mathematical expressions that combine linear and non-linear forms on complexes. For instance, it makes sense to solve, or train, an expression of the form $d_1(TN_{d_0}(f) )= L_2(g)$, where $f$ a cochain in $\mathcal{C}^0$ and $g$ is a given cochain in $\mathcal{C}^2$. Second, this form generalizes (with the right choice of the matrix $\mathcal{A}$) all existing message passing protocols defined on complexes including \cite{hajijcell,roddenberry2021principled,schaub2021signal,schaub2020random,bick2021higher,ebli2020simplicial}, and hence offering a unifying framework to write deep learning on graphs, surfaces, and higher order complexes. Note that we are defining $\mathcal{A}$ to operate on a cochain defined on the entire complex $\mathcal{M}$. This means that the message passing can occur among simplices of different dimensions simultaneously. In practice, the operator $\mathcal{A}$ is a block matrix whose block elements are the discrete operators we introduced in Section 3.1.1 or operators that can be built from them (e.g. the exterior derivative, the Laplacian, etc.). Third, from a practical standpoint, this setting creates a concrete mathematical foundation for a unifying deep learning API on complexes as well as DEC. This can ultimately invite practitioners from both fields to push the state-of-the-arts in new directions. With the right abstraction of data, we envision a unifying API across multiple algorithms and disciplines related to data science.

\vspace{-8pt}

\section{Related Works}
\vspace{-4pt}
\label{related}
 Spivak et al. \cite{spivak2012ologs} introduced a categorical framework for knowledge representation. Another effort that focuses on creating concrete foundations for geometric deep learning is proposed by Bronstein et al. \cite{bronstein2021geometric}. Our work approaches the notion of data in general terms instead of focusing on deep models in particular. Moreover, our work emphasizes the role re-examining data notion plays in understanding models from a theoretical and practical standpoints. Recent works on non-linear graph signal processing and its generalization to higher order complexes \cite{roddenberry2021principled,hajijcell,hajij2021simplicial,bick2021higher,schaub2021signal, roddenberry2019hodgenet} implicitly discuss the cochain notion of data. In \cite{schaub2021signal}, Schaub et al. emphasized the relationship between linear signal processing on complex (determined by linear operators) and non-linear filtering defined via a neural network on higher order networks. The goal of these previous works is to study signal processing on higher order complexes. Our goal here is to emphasize the unifying principles offered by the categorical and cochain notions and its importance in the data-centric AI transition. In particular, we emphasize the theoretical advantages, practical ramifications, and the evolution of general understanding of these principles.      
\vspace{-6pt}
\section{Acknowledgement}

Mustafa Hajij was supported in part by the National Science Foundation (NSF, DMS-2134231).

\bibliographystyle{abbrv}

\bibliography{refs}

\end{document}